\documentclass[]{articulo}

\title{LiveChess2FEN: a Framework for Classifying Chess Pieces based on CNNs}
\author{David Mallasén Quintana \and Alberto Antonio del Barrio García \and Manuel Prieto Matías}

\begin{document}

\maketitle

\begin{abstract}
Automatic digitization of chess games using computer vision is a significant technological challenge. This problem is of much interest for tournament organizers and amateur or professional players to broadcast their over-the-board (OTB) games online or analyze them using chess engines. Previous work has shown promising results, but the recognition accuracy and the latency of state-of-the-art techniques still need further enhancements to allow their practical and affordable deployment. We have investigated how to implement them on an Nvidia Jetson Nano single-board computer effectively. Our first contribution has been accelerating the chessboard's detection algorithm. Subsequently, we have analyzed different Convolutional Neural Networks for chess piece classification and how to map them efficiently on our embedded platform. Notably, we have implemented a functional framework that automatically digitizes a chess position from an image in less than $1$ second, with $92$\% accuracy when classifying the pieces and $95$\% when detecting the board.
\end{abstract}

\section{Introduction}

The recent breakthroughs in Deep Neural Networks \cite{Lecun1998, Krizhevsky2012, He2016, Sze2017} have provided an astonishing advance in the application of image classification algorithms. Many Convolutional Neural Networks (CNNs) have been employed for this task, such as MobileNet \cite{Sandler2018}, Xception \cite{Chollet2016} or NASNet \cite{Zoph2018}. These CNNs have been continually studied and improved to perform well even for large-scale image classification \cite{Krizhevsky2012}.

Nevertheless, the efficient recognition of chess pieces and chessboards is still an unsolved computer vision problem \cite{Czyzewski2017, Delgado2019}. Its solution will benefit experienced players who want to study and train using chess engines and other specialized software programs but prefer to work with a physical chessboard. It will also be useful for chess tournament organizers who want to broadcast OTB games online or for amateur players who want to share their OTB games with friends.

Specialized chess sets (electronic boards) can efficiently perform this digitization task, but they are expensive. As an alternative, previous work has explored affordable solutions based exclusively on computer-vision \cite{Czyzewski2017, Ding2016}. Overall, the problem can be broken down into two main parts. The first step is to recognize the chessboard and its orientation, and then identify the chess pieces and their precise position afterwards.

In this paper, we present LiveChess2FEN, a framework that tries to fulfill the expectations raised by that state-of-the-art solutions \cite{Czyzewski2017, Ding2016}. We have focused on optimizing the recognition process and reduce its latency as much as possible. Note that in addition to accuracy, some of the envisioned applications require very low latencies. This is the case of broadcasting OTB games online where players can make moves even in fractions of a second, especially in game modes known as ``Bullet'' or ``Blitz''.

Remarkably, LiveChess2FEN can identify all the chess pieces of a given position in less than a second, outperforming the recognition latency of previous work by a factor of five \cite{Czyzewski2017}. This improvement has been possible thanks to the deployment of CNNs on top of an Nvidia Jetson Nano single-board computer and the implementation of additional optimizations that exploit domain-specific information related to the chess rules. In this manner, we avoid the use of computationally-expensive chess engines such as Stockfish \cite{Silver2017}.

The rest of the paper is organized as follows: Section \ref{sec:related work} describes the state of the art in chessboard and pieces recognition. Section \ref{sec:preliminaries} introduces FEN notation, the final output of the LiveChess2FEN framework, which is described in Section \ref{sec:LiveChess2FEN}. Section \ref{sec:optimizations_acceleration} details the chosen hardware platform and the optimizations made. Section \ref{sec:experiments_results} reports the results of each of the digitization steps and the performance achieved for the full digitization. Finally, Section \ref{sec:conclusions} outlines the conclusions and future lines of work.

\section{Related work}
\label{sec:related work}

Specialized boards that physically detect the pieces are a hardware solution for automatic chess digitization \cite{SquareOff2019,DGTech}. However, these boards are expensive and difficult to deploy in many areas. For example, a set of DGT chessboard and pieces (brand used in official tournaments) costs between \euro500 and \euro1000 \cite{DGTechShop}.

Other alternatives are those provided by robots that move the pieces on a board, such as \cite{Matuszek2011} or more recently \cite{Chen2019} and \cite{Kolosowski2020}. These robots are based on positioning a zenith camera over the board and detecting the differentials between one movement and the next. A drawback of this approach is that it is necessary to start from a known initial position. A board with a generic position could not be digitized this way. In addition, errors should be taken into account because they could add up in each new play.

The solutions offered by computer vision are an alternative to consider since they provide cheaper and increasingly accurate systems. Furthermore, there are several platforms with enough computational power to perform these tasks at an affordable cost. For instance, the Nvidia Jetson Nano costs around \euro110 \cite{JetsonNano} and does not just stick to one function, it could be reused for other tasks. Other alternatives include the Intel Neural Compute Stick 2 \cite{Mas2020} or Google's Coral boards \cite{GoogleCoral}.

These computer vision procedures are based on combining and adapting transformations and detectors already known and used in other fields, such as the Harris corner detector and the Hough transform \cite{Escalera2010}. Many of the methods often assume significant simplifications, such as determining the exact position of the camera, using boards designed explicitly with markers to aid in corner detection, or directly through user interaction \cite{Ding2016}. However, some generic solutions that overcome these restrictions already exist.

For example, some methods allow us to classify occurrences of various objects in arbitrary places using CNNs \cite{Gao2017}. However, they do not have the precision required to obtain their exact location. The authors of \cite{Bency2016} describe a method for object detection that also uses CNNs trained through weak supervision. That is, it is enough to have labeled images of the objects to train the network, without the need to provide the exact location of the object in each training image. The problem with this approach is that it takes many iterations to locate an object precisely, which does not make it very practical in situations that require fast responses.

The authors of \cite{Czyzewski2017} propose a method for chessboard detection that is robust against light conditions and the angle from which the images are taken. In addition, it works with most styles of boards and it overcomes many of the weaknesses that we have been discussing. It is an iterative process in which the location of the board is refined in several phases. The authors obtain $99.5$\% accuracy in detecting the intersections of the center board grid and find the full location of the chessboard accurately $95$\% of the time.

Regarding piece classification, once the board has been located, in \cite{Ding2016}, a method is proposed that is based on Support Vector Machine (SVM). This is trained on the features extracted by SIFT \cite{Lowe2004}, thus achieving $85$\% accuracy when classifying the pieces. In \cite{Delgado2019} a method for training CNNs from artificially generated 3D images is introduced. Authors obtain $97$\% accuracy, although these are computer-generated scenarios, and no testing with real chessboard and pieces is done.

In \cite{Xie2018} authors introduce an alternative to CNNs, oriented chamfer matching. They obtain comparable results to those of CNNs, but with a lower training set in exchange for fixing the piece types, as they rely heavily on template matching. In \cite{Czyzewski2017} authors claim to achieve $95$\% accuracy classifying chess pieces using a custom CNN, which they enhance by clustering similar pieces, taking into account their height and area, and using a game engine (Stockfish) to obtain the probability of particular positions. The code for these enhancements has not been released, so we could not analyze this particular method.

In this paper, we leverage the approach in \cite{Czyzewski2017} to detect the board. Several optimizations have been performed to accelerate its execution, as described above. As for the piece classification in an arbitrary snapshot, different CNNs have been studied and mapped onto an Nvidia Jetson Nano board. Finally, all the different scenarios are tested and put together to form the LiveChess2FEN framework.

\section{Preliminaries: FEN notation}
\label{sec:preliminaries}

FEN notation \cite{Edwards1994} is a string representation of a position of the board. This will be the final output of our digitization, as it can be imported directly into chess engines or other computer programs to visualize a digital chessboard.

Since only a snapshot of the game will be available, only the pieces' position at a particular moment in time will be known. Not the player whose turn it is to move or if there is a possibility of castling for example (factors that are taken into account in the full FEN standard). Thus, the output string is a series of eight blocks of alphanumeric characters representing each row of the board separated by the / character. For instance, the initial position of a game is encoded as: rnbqkbnr/pppppppp/8/8/8/8/PPPPPPPP/RNBQKBNR.

\section{LiveChess2FEN} \label{sec:LiveChess2FEN}

\begin{figure}[]
	\centering
	\setlength{\fboxsep}{0pt}
	\setlength{\fboxrule}{1pt}
	\fbox{\includegraphics[width=0.75\textwidth]{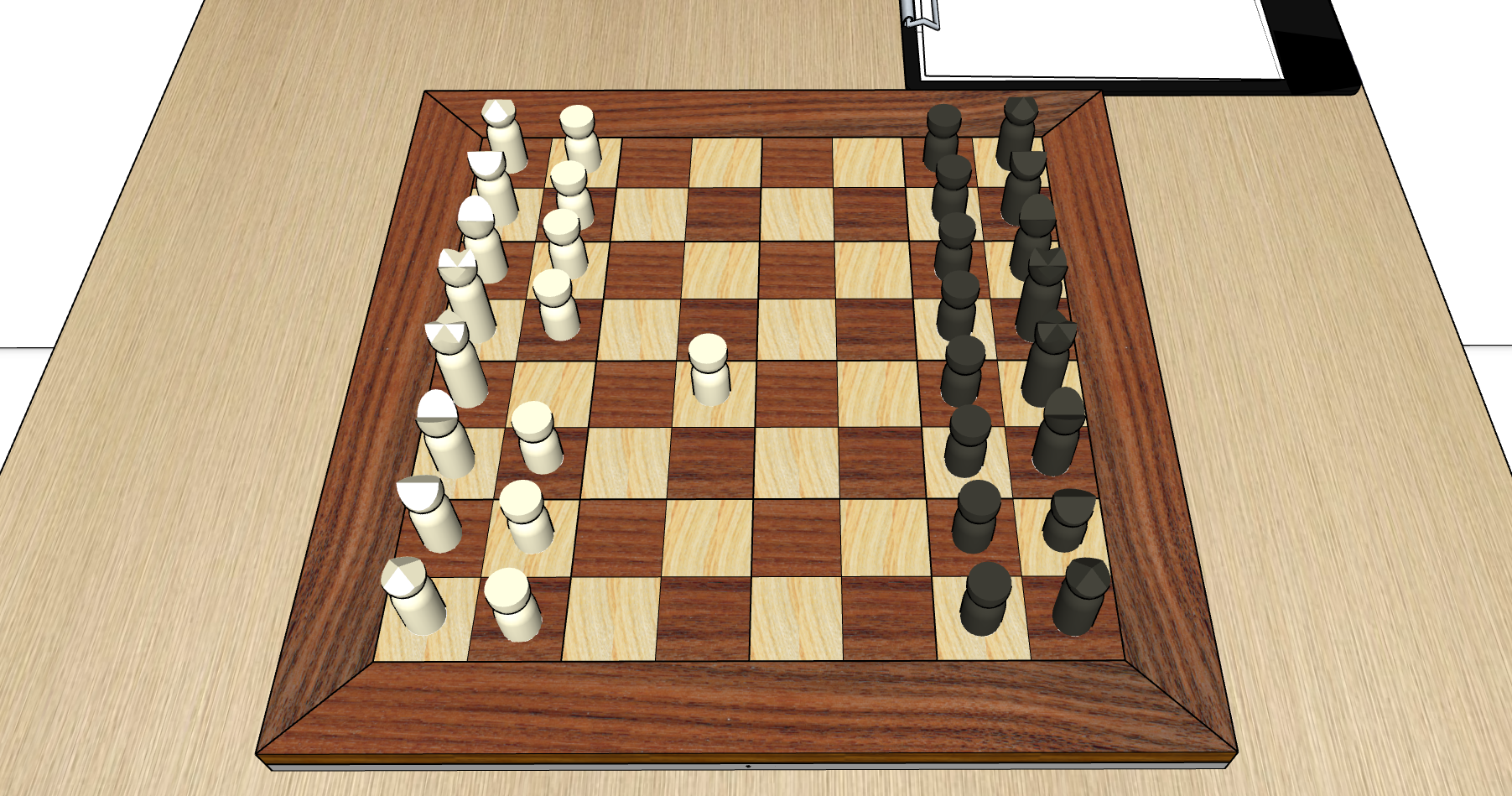}}
	\caption{An example of a photo taken by the camera.}
	\label{fig:vision_camara}
\end{figure}

\begin{figure}[]
	\centering
	\includegraphics[width=0.8\textwidth]{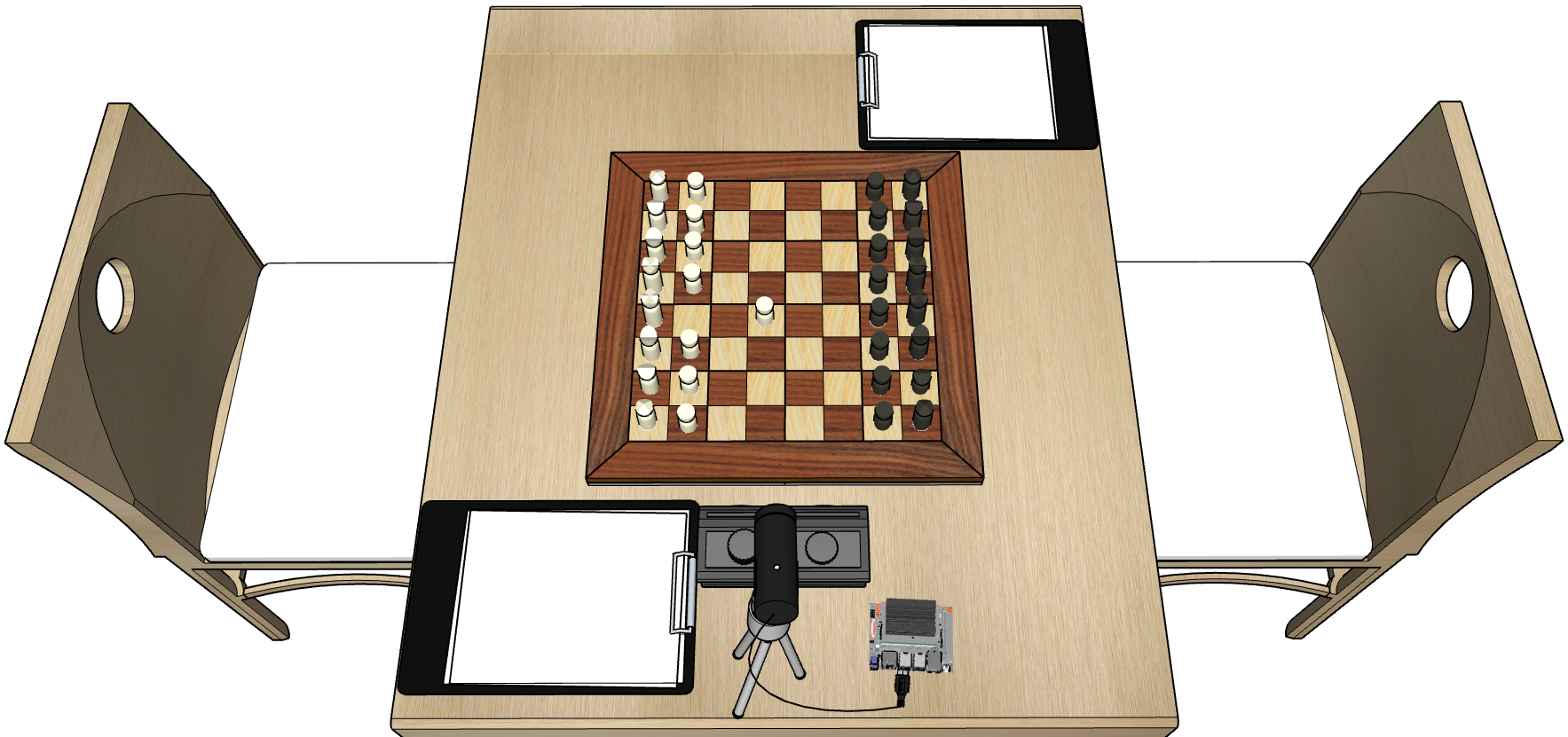}
	\caption{Schematic of the camera position and the specialized hardware that will execute the digitization process.}
	\label{fig:posicion_camara}
\end{figure}

LiveChess2FEN is a functional framework capable of executing the entire process of digitizing a chess game photo in real-time, making all the necessary calculations on specialized hardware. This has been carried out having in mind its use in amateur games or in tournaments, taking the photos from the side of the board each time a player presses the clock (Figure \ref{fig:vision_camara}). We show a schematic of the camera position in Figure \ref{fig:posicion_camara}.

\begin{figure}[]
    \centering
    \includegraphics[width=0.9\textwidth]{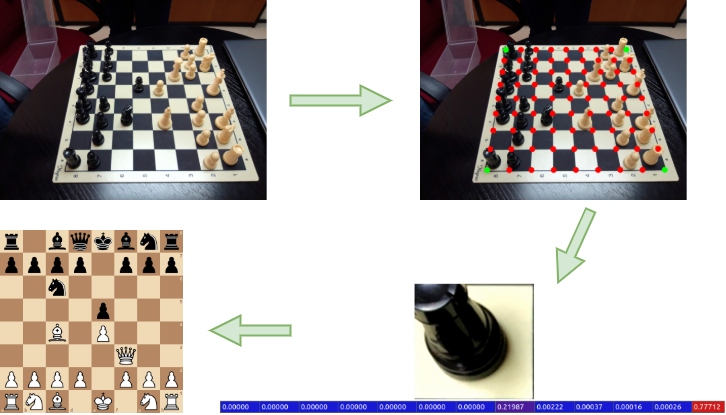}
    \caption{Full digitization process.}
    \label{fig:digitization_process}
\end{figure}

The full digitization process is done in two major steps. The first step is to locate the chessboard in the input image. Subsequently, the algorithm must classify each of the squares into the corresponding chess piece. An overview of the full process is shown in Figure \ref{fig:digitization_process}.

\subsection{Board detection} \label{subsec:LC2FEN_board_detection}

Locating a chessboard with pieces hiding part of its grid is a complex machine vision problem. Nevertheless, this step must be extremely precise when locating the board's four corners since these coordinates are used to split it into its 64 squares. With this information, we can crop the empty squares and the chess pieces from the original image and locate their positions automatically. Reducing errors in this step is essential to classify the chess pieces correctly afterward. Authors propose in \cite{Czyzewski2017} a fast iterative process, which starting from a photo taken from the surroundings of a board, is able to locate it with enough precision after just a few iterations. Our proposal is based on this previous work and introduces further optimizations to reduce the execution cost and thus the latency, as shown in Section \ref{sec:optimizations_acceleration}.

In situations where the board and camera are going to stay still, the framework can take advantage of this fact to avoid recalculating the location of the board each time. In these cases, it must store the coordinates of the corners calculated for the previous image, check that the board is still in the same place and proceed to separate the squares directly.

We have implemented an algorithm that can quickly check if the board is still in the same location. For this, the algorithm employs the geometric detector and the neural network used when locating the chessboard grid points in the board detection phase. If the 49 corners of the central $6 \times 6$ square of the board are still in place, the previous information that is stored is still correct. When counting the corners that are part of the grid, a tolerance margin must be taken since, for instance, some points may be occluded by a piece. Experimenting on different test images, we have concluded that spotting 20 of the 49 points is enough to confirm that the board is still in the same place (Algorithm \ref{alg:comprobacion_posicion_tablero}).

\begin{algorithm}[]
    \caption{Board location check.}
    \label{alg:comprobacion_posicion_tablero}

    \SetKwData{lineas}{lines}
    \SetKwData{imagen}{image}
    \SetKwData{punto}{point}
    \SetKwData{puntoscuadricula}{grid\_points}
    \SetKwData{matriz}{matrix}
    \SetKwData{tolerancia}{tolerance}
    \SetKwData{esquinas}{corners}
    \SetKwData{esquinascorrectas}{correct\_corners}
    \SetKwData{espuntocuadricula}{is\_grid\_corner}
    \SetKwFunction{intersecciones}{intersections}
    \SetKwFunction{vecindad}{neighborhood}
    \SetKwFunction{preprocesar}{preprocess}
    \SetKwFunction{append}{append}
    \SetKwFunction{detectorgeometrico}{geometric\_detector}
    \SetKwFunction{redneuronal}{neural\_net}
    \KwIn{Image containing a board.\newline Candidate corners to continue being part of the central $6\times 6$ square of the board.}
    \KwOut{If the board is still in the same location.}
    \BlankLine
    \esquinascorrectas $\leftarrow$ 0\;
    \ForEach{\punto $\in$ \esquinas} {
        \matriz $\leftarrow$  \preprocesar{\vecindad{\imagen, \punto}}\;
        \espuntocuadricula $\leftarrow$ \detectorgeometrico{\matriz}\;
        \eIf{\espuntocuadricula} {
            \esquinascorrectas $\leftarrow$ \esquinascorrectas + 1\;
        }{
            \espuntocuadricula $\leftarrow$ \redneuronal{\matriz}\;
            \If{\espuntocuadricula} {
                \esquinascorrectas $\leftarrow$ \esquinascorrectas + 1\;
            }
            \tcp{If it does not belong to the grid, go on to the next one}
        }
    }
    \KwRet{\esquinascorrectas $\geq$ \tolerancia $(= 20)$}
\end{algorithm}

\subsection{Chess piece classification} \label{subsec:piece_classification}

Once the board's location in the image is known, it must be separated into its individual squares to classify them and obtain the final result. This is done by simply dividing the image from the last iteration of the chessboard detection, a square, into an $8 \times 8$ grid.

After dividing the original image into each of the board's squares, the next step is to classify them. Each square can be empty or occupied by a piece of one of the players, so it is necessary to decide which of the 13 classes corresponds to each of the board's 64 squares. Currently, the most widespread and the best algorithms to classify an image into a series of categories are CNNs \cite{Rawat2017}. Furthermore, these networks can be significantly accelerated, as shown in Section \ref{sec:optimizations_acceleration}.

The final result of the classification is a string that encodes using FEN notation the position extracted from the image, as described in Section \ref{sec:preliminaries}.

In order to train a deep learning model, a large number of images are necessary. Furthermore, if the network must also classify different types of pieces, it is necessary to introduce this variety in the training data. To train our models with enough variety, two labeled datasets of chess pieces \cite {Yang2016} \cite{Schubiner2019} have been put together. Thus, the employed dataset consists of nearly $55000$ images of chess pieces.

The high-level Keras API \cite{Chollet2015} on top of the TensorFlow library \cite{Abadi2015} was used to define and train the CNNs with which to test the piece classification. Transfer learning and fine-tuning techniques were used to ease this process. To choose the pre-trained models for our dataset, we leveraged the exhaustive study carried out in \cite{Bianco2018}. This was especially useful since the authors also tested different configuration options available on the Nvidia Jetson Nano, which has also been our platform of choice for the inference.

Based on these results, we decided to test MobileNetV2 \cite{Sandler2018}, NASNetMobile \cite{Zoph2018}, DenseNet201 \cite{Huang2017}, Xception \cite{Chollet2016}, AlexNet \cite{Krizhevsky2012} and SqueezeNet-v1.1 \cite{Iandola2016}. AlexNet and SqueezeNet-v1.1 have the lowest inference times, especially when executing in batches. MobileNetV2 is a larger model, but it is very versatile, as it can be tuned with an $\alpha$ parameter to control the width of the network. NASNetMobile, DenseNet201 and Xception are more complex and thus slower, but achieve a higher accuracy while still requiring a reasonable amount of resources to run.

The output of the deep learning models is a 13 component vector. Besides using this vector of probabilities, some domain knowledge has been integrated into our flow to improve the piece classification accuracy. In this way, chess rules were introduced to take into account all of the squares at the same time, as opposed to the classification of each square using just the output of the deep learning model, which is agnostic of its surroundings. This process is summarized in Algorithm \ref{alg:inferencia}.

\begin{algorithm}[]
	\caption{Calculation of the position from the probability vectors of each square.}
	\label{alg:inferencia}

	\SetKwData{tablero}{board}
	\SetKwData{tops}{tops}
	\SetKwData{vectoresprobs}{prob\_vectors}
	\SetKwData{reyblanco}{white\_king}
	\SetKwData{casillavacia}{empty\_square}
	\SetKwData{casilla}{square}
	\SetKwData{probcasilla}{square\_prob}
	\SetKwData{porrellenar}{to\_fill}
	\SetKwData{pieza}{piece}
	\SetKwData{piezasusadas}{used\_pieces}
	\SetKwFunction{maxprob}{max\_prob}
	\SetKwFunction{alcanzadomax}{max\_reached}
	\SetKwFunction{sort}{sort}

	\KwIn{Probability vectors of each square.}
	\KwOut{Board position.}
	\BlankLine

	\tablero $\leftarrow$ $[\ ] * 64$ \tcp{Empty list of the 64 squares}
	\BlankLine

	\tcp{Set the kings, equally with the black king}
	\reyblanco $\leftarrow$ \maxprob{\vectoresprobs, \texttt{`K'}}\;
	$\tablero[\text{\reyblanco}]\ \leftarrow$ \texttt{`K'}\;
	\dots\\
	\porrellenar $\leftarrow 62$\;
	\BlankLine

	\tcp{Set the empty squares}
	\ForEach{\casilla $\in$ \vectoresprobs} {
		\If{\maxprob{\casilla} = \texttt{`\_'}}{
			$\tablero[\text{\casilla}]\ \leftarrow$ \texttt{`\_'}\;
			\porrellenar $\leftarrow$ \porrellenar $-\ 1$\;
		}
	}
	\BlankLine

	\tcp{Sort the probability vectors obtaining the lists shown in Figure \ref{fig:algoritmo_config_tablero} and the \texttt{tops} vector}
	\dots
	\BlankLine

	\tcp{Finish filling up the board in the order given by the piece probabilities}
	\While{\porrellenar $> 0$}{
		\pieza $\leftarrow$ \maxprob{\tops}\;
		\If{$\neg$\alcanzadomax{\pieza, \piezasusadas} $\land\ \tablero[\pieza] =[\ ]$}{
			$\tablero[\pieza]\ \leftarrow$ \pieza\;
			\porrellenar $\leftarrow$ \porrellenar $-\ 1$\;
			$\piezasusadas[\pieza] \leftarrow \piezasusadas[\pieza] + 1$\;
		}
		\tcp{Update the lists and the pointers of \texttt{tops}}
		$\tops[\pieza] \leftarrow \emptyset$\;
		\dots
	}
	\KwRet{\tablero}
\end{algorithm}

Firstly, the algorithm finds the two squares with the greatest probability of containing the kings. This ensures that there is exactly one king of each color. Subsequently, all of the empty squares are set since the models detect them with enormous precision. Finally, the rest of the squares are classified, considering that each piece has a maximum number of appearances. In addition, if a player keeps the bishop pair, those bishops must be in squares of different colors \footnote{In our study, we have assumed that, if there were any previous untracked promotions, they have been to a queen. In a promotion, the player chooses any piece except a king or another pawn. However, the occasions when a player does not choose a queen are rare. This assumption was made since no knowledge of previous positions is expected, but it can be completely eliminated if the whole game's history is available. In these cases, it would be possible to precisely know the number and type of pieces each player has.}.

\begin{figure}[]
	\centering
	\includegraphics[width=\textwidth]{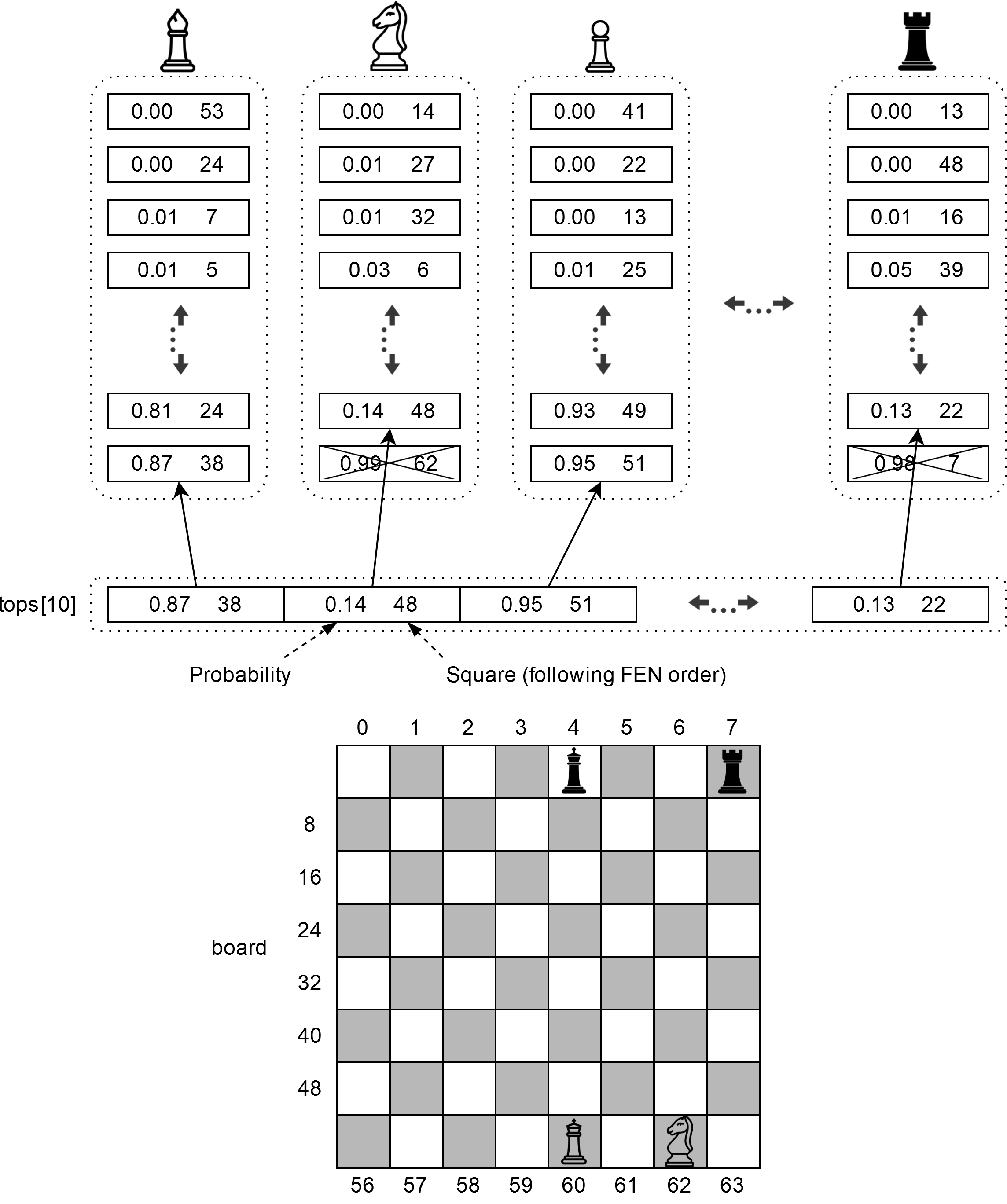}
	\caption{Outline of the data used by the inference of the position on the board. The kings are set at the beginning of the algorithm. In this snapshot, also a white knight, which had a probability of $0.99$ of being in square $62$, and a black rook, which had a probability of $0.98$ of being in square $7$, have been set.} \label{fig:algoritmo_config_tablero}
\end{figure}

In order to classify the pieces that are not kings, the program follows the following steps. First, the remaining squares are sorted according to their probability of containing each of the remaining $10$ classes (in total, there were $13$, but the kings' positions and the empty squares have already been decided). Thus, $10$ lists of pairs of pieces and squares are obtained ordered from highest to lowest probability of containing the corresponding piece (the top vertical lists in Figure \ref{fig:algoritmo_config_tablero}). Afterward, the algorithm iterates by choosing the element at the top of the list with a higher probability (\texttt{tops} vector). If the maximum occurrences of that type of piece have not yet been reached and the square it represents on the board has not yet been filled, it is selected. In any case, the piece is removed from its list of ordered pieces, as it has already been processed (Figure \ref{fig:algoritmo_config_tablero}).

In the next iteration, the piece that has the second-highest probability is chosen, and the process is repeated. The algorithm finishes after covering the entire board. In this way, as shown in Section \ref{sec:experiments_results}, the piece classification accuracy has been increased.

\section{Optimizations and acceleration} \label{sec:optimizations_acceleration}

The final step and the main contribution of our work has been accelerating the entire framework on a target embedded platform. Our platform of choice is the popular  Nvidia Jetson Nano single-board computer \cite{JetsonNano}, which offers 472 GFLOPs of FP16 compute performance.

In addition to a dedicated Nvidia GPU, this system integrates a quad-core ARM CPU capable of performing the sequential computation necessary to detect the chessboards. This type of architecture has been widely used successfully for more than a decade in tasks related to image processing \cite{Setoain2007} \cite{Tenllado2008}.

\begin{figure}[]
    \centering
    \includegraphics[width=\textwidth]{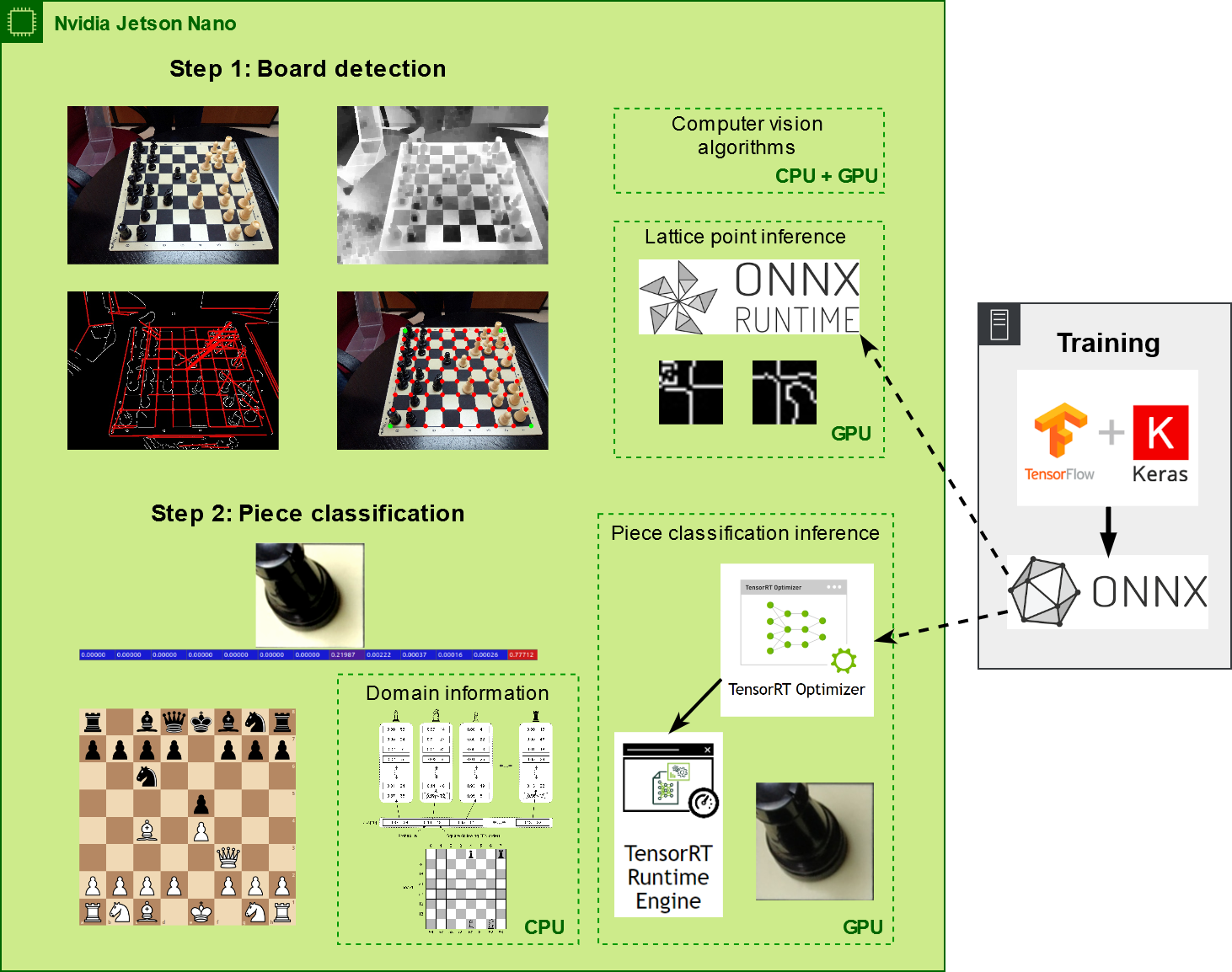}
    \caption{Workflow for the full digitization.}
    \label{fig:full_workflow}
\end{figure}

The neural networks have been defined and trained using Keras on top of TensorFlow. In order to improve the inference execution times, we have leveraged the ONNX (Open Neural Network Exchange) model representation format \cite{ONNX} and its optimizer ONNXRuntime, \cite{ONNXRuntime} as well as TensorRT \cite{TensorRT}, the deep learning inference optimizer developed by Nvidia (Figure \ref{fig:full_workflow}).

The execution times of the whole framework have been further reduced by optimizing the bottlenecks spotted in the original flow when detecting the board \cite{Czyzewski2017}. The most important improvements were the following:
\begin{itemize}
    \item Accelerating the inference of the neural network used to classify the most complex lattice points of the board's grid. For this purpose, the original model was transformed into the ONNX format, and its execution was optimized with ONNXRuntime.
    \item Reducing the overhead caused by the calls to the NumPy library \cite{Oliphant2006}. When computing the distance from a point to a line in two dimensions, the general formula $\norm{(y - x) \times (x - z)}$ can be simplified to $$\abs{(y_1 - x_1)(x_2 - z_2) - (y_2 - x_2)(x_1 - z_1)},$$ which can be calculated without the need of complex mathematical libraries.
    \item Simplifying some calculations when computing the intersections of a set of lines. In cases where the size $n$ of the set is small, the Bentley-Ottmann algorithm \cite{Bentley1979}, which has an asymptotic complexity of $\mathcal{O}((n+k)\log n)$, where $k$ is the number of intersections, is actually significantly slower than the naive approach, which has an asymptotic complexity of $\mathcal{O}(n^2)$.
\end{itemize}

\section{Experiments and results} \label{sec:experiments_results}

In this section, the final results are analyzed and compared with the initial results and with other proposals. In addition, the total time that the program takes to execute the complete digitization is studied. Starting by loading the photo of a board and finishing with the calculation of the FEN notation of its position.

\subsection{Board detection}

The modifications introduced in the baseline code \cite{Czyzewski2017} do not affect the final precision when detecting the board. Therefore, our flow maintains the same accuracy: $99.5$\% when detecting the intersections of the board's central grid and accurately finds the location of the board in the image $95$\% of the times.

Regarding the final execution times, a noteworthy improvement has been achieved. The set of 10 photos used in \cite{Czyzewski2017} has been employed in order to calculate the runtimes. As the goal is to minimize the latency, we measure the time elapsed since the beginning of the loading of the input photo and until the cropped board image is saved. All the optimizations mentioned in Section \ref{sec:optimizations_acceleration} have reduced the latency in detecting the board by a factor of four. A speedup of $4.27$ has been achieved, reducing the time required from $16.38$ to $3.84$ seconds per board on average (Table  \ref{tab:tiempos_tablero}).

\begin{table}[h]
	\centering
		\begin{tabular}{@{}lcccccc@{}}
			\toprule
			& Initial & Adapted & ONNX     & NumPy   & Bentley-Ottmann & Final\, \\ \midrule
			\, Time  & $16.38$s & $16.01$s & $10.33$s & $5.55$s & $4.22$s & $\mathbf{3.84}$\textbf{s}\,     \\
			\textit{\, Speedup} & -        & $1.02$      & $1.55$   & $1.86$  & $1.32$          & $1.10$\,     \\
			\, Accumulated& -        & $1.02$      & $1.59$   & $2.95$  & $3.88$          & $\mathbf{4.27}$\,     \\ \bottomrule
		\end{tabular}%
	\caption{Average time per image on the Jetson Nano for each of the board detection optimizations. Speedups are obtained with respect to the previous step and the accumulated speedup with respect to the initial time.}
	\label{tab:tiempos_tablero}
\end{table}

It must also be considered that in \cite{Czyzewski2017}, authors comment that they get much higher accuracy in exchange for an execution time increase, which in the end is up to two times slower than other alternatives. Hence, we could conjecture that this improved method is twice as fast as those proposed in \cite{Escalera2010} and \cite{Danner2015}, as well as being more accurate.

\subsection{Piece classification}

In this section, we have evaluated the accuracy when classifying pieces by means of the \textit{Top-1} value after applying the corresponding CNN model. Moreover, this has been complemented by including the domain-knowledge improvements. To gather these measurements, tests on 5 chessboard photos\footnote{The test photos can be downloaded from \url{https://github.com/davidmallasen/LiveChess2FEN/releases/tag/v0.1.0}} have been carried out. Each board has between 21 and 32 pieces in various positions drawn from real games.

To avoid overfitting when evaluating the accuracy, these boards contain pieces of a different type from the ones used in the training dataset. In this way, the robustness of each model to changes in the type of the pieces can be verified.

\begin{table}[]
    \centering
    \resizebox{\textwidth}{!}{%
        \begin{tabular}{l|cc|cc|cc|cc}
            & \multicolumn{2}{c|}{Xception} & \multicolumn{2}{c|}{DenseNet201} & \multicolumn{2}{c|}{NASNetMobile} & \multicolumn{2}{c}{MobileNetV2} \\ \hline
            Test 1 & 95\% & 95\% & 91\% & 94\% & 94\% & 94\% & 95\% & 98\% \\
            Test 2 & 92\% & 94\% & 86\% & 95\% & 91\% & 92\% & 89\% & 89\% \\
            Test 3 & 97\% & 95\% & 94\% & 94\% & 91\% & 91\% & 92\% & 92\% \\
            Test 4 & 89\% & 91\% & 91\% & 91\% & 89\% & 91\% & 89\% & 91\% \\
            Test 5 & 91\% & 97\% & 83\% & 88\% & 97\% & 98\% & 92\% & 92\% \\ \hline
            Media  & 93\% & \textbf{94\%} & 89\% & \textbf{92\%} & 92\% & \textbf{93\%} & 91\% & \textbf{92\%} \\ \hline
            & \textit{Top-1} & Domain & \textit{Top-1} & Domain & \textit{Top-1} & Domain & \textit{Top-1} & Domain \\
        \end{tabular}
    }
    \\\vspace{0.5cm}
    \hspace{0.1cm}
    \resizebox{\textwidth}{!}{%
        \begin{tabular}{l|cc|cc|cc|cc}
            &
            \multicolumn{2}{c|}{MobileNetV2} &
            \multicolumn{2}{c|}{MobileNetV2} &
            \multicolumn{2}{c|}{\multirow{2}{*}{AlexNet}} &
            \multicolumn{2}{c}{\multirow{2}{*}{SqueezeNet-v1.1}} \\
            &
            \multicolumn{2}{c|}{$\alpha=0.5$} &
            \multicolumn{2}{c|}{$\alpha=0.35$} &
            \multicolumn{2}{c|}{} &
            \multicolumn{2}{c}{} \\ \hline
            Test 1 & 95\% & 95\% & 84\% & 89\% & 83\% & 83\% & 91\% & 97\% \\
            Test 2 & 86\% & 91\% & 72\% & 75\% & 61\% & 61\% & 81\% & 86\% \\
            Test 3 & 94\% & 94\% & 80\% & 83\% & 77\% & 80\% & 89\% & 92\% \\
            Test 4 & 84\% & 88\% & 81\% & 83\% & 73\% & 78\% & 83\% & 84\% \\
            Test 5 & 89\% & 91\% & 86\% & 89\% & 72\% & 72\% & 89\% & 94\% \\ \hline
            Media  & 90\% & \textbf{92\%} & 81\% & \textbf{84\%} & 73\% & \textbf{75\%} & 87\% & \textbf{91\%} \\ \hline
            & \textit{Top-1} & Domain & \textit{Top-1} & Domain & \textit{Top-1} & Domain & \textit{Top-1} & Domain \\
        \end{tabular}
    }
    \caption{\textit{Top-1} value and accuracy after including the global domain knowledge into the inference of each of the models.}
    \label{tab:jetson_top1vsdominio}
\end{table}

These results are shown in Table \ref{tab:jetson_top1vsdominio}. It is possible to observe that the accuracy increases between $ 1 $ and $ 4 $ \% on average when including domain information. However, it must be noted that, in some cases, including this knowledge worsens the accuracy slightly due to corner cases. For example, if there are two squares with a very high probability of containing a black king, it may be the case that this piece is located in the square that has a slightly lower probability than the other one. In this case, the domain information algorithm would choose the wrong square to position the king, but the probability vectors would assign a black king to both squares. Therefore, the former algorithm would fail to guess both of the squares, and the latter would guess correctly one of them.

Nevertheless, this global domain knowledge increases the accuracy in most situations and inserts coherence in the results. At all times, they are positions that could occur in a real chess game. Finally, it must be noted that these accuracies remain similar when executing the piece classification models on the optimizers that have been considered.

The execution time is depicted in Table \ref{tab:resumen_modelos}. In this table, the best results achieved with each model are displayed. As can be seen, the best inference engine has always been TensorRT with a batch size of 64, except in the NASNetMobile case, where it was not possible to transform the ONNX model. Notably, the original inference times of the models ranged between $ 5.86 $ and $ 28.96 $ seconds per board, while in our case, these range from $ 0.46 $ to $ 6.04 $ seconds.

\begin{table}[]
    \centering
    \begin{tabular}{@{}lccc@{}}
        \toprule
        & Average time & Average accuracy & Inference engine \, \\ \midrule
        \, SqueezeNet-v1.1 & $0.46$s      & 91\%            & TensorRT b64        \\ \midrule
        \begin{tabular}[c]{@{}l@{}}\, MobileNetV2\\\, $\alpha=0.35$\end{tabular} & $0.52$s & 84\% & TensorRT b64 \\ \midrule
        \begin{tabular}[c]{@{}l@{}}\, MobileNetV2\\\, $\alpha=0.5$\end{tabular}  & $0.60$s & 92\% & TensorRT b64 \\ \midrule
        \, AlexNet         & $0.62$s      & 75\%            & TensorRT b64        \\ \midrule
        \, MobileNetV2     & $0.90$s      & 92\%            & TensorRT b64        \\ \midrule
        \, NASNetMobile    & $3.42$s      & 93\%            & ONNXRuntime         \\ \midrule
        \, Xception        & $5.62$s      & 94\%            & TensorRT b64        \\ \midrule
        \, DenseNet201     & $6.04$s      & 92\%            & TensorRT b64        \\ \bottomrule
    \end{tabular}
    \caption{Best execution times per board and accuracies obtained for each model on the Jetson Nano.}
    \label{tab:resumen_modelos}
\end{table}

\begin{figure}[]
    \centering
    \includegraphics[width=0.8\textwidth]{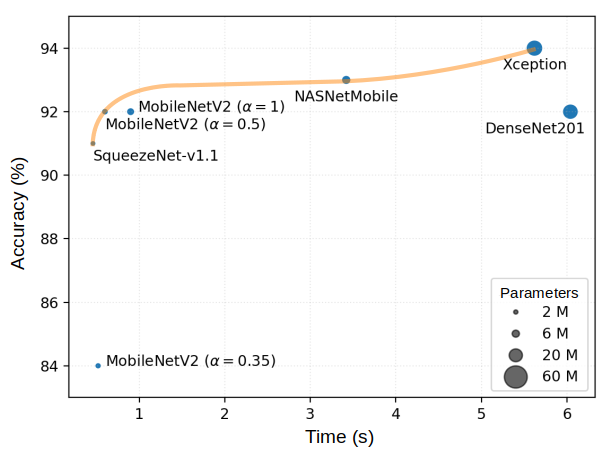}
    \caption{Pareto Front when studying the accuracy and execution time of each model.}
    \label{fig:grafico_tiempoprecision}
\end{figure}

According to Table \ref{tab:resumen_modelos} there is no best solution, so Figure \ref{fig:grafico_tiempoprecision} shows the Pareto Front when studying both accuracy and execution time for the aforementioned models and optimizers. Furthermore, the number of parameters of each network is illustrated through circles of different sizes. Four models appear on the Pareto Front, namely: SqueezeNet-v1.1, MobileNetV2($\alpha=0.5$), NASNetMobile and Xception. Among these, Xception and NASNetMobile possess the highest accuracies. However, they require more than 5 and 3 seconds, respectively, to perform the classification of all the squares of the board in the image. Hence, for the scope of this work, SqueezeNet-v1.1 and MobileNetV2($\alpha=0.5$) are the best candidates, as their execution time is far below 1s and their accuracy exceeds 90\%.

\subsection{Full digitization}

The full digitization of a chessboard comprises from the moment the system detects that there is a new image to be processed until it finishes processing it and produces the output FEN string. Besides the detection of the board in the initial image and the classification of the pieces, which practically involve the entire execution time, three additional operations are required to complete the whole process.

Firstly, once the chessboard has been detected, the individual squares must be separated. This takes around $80$ milliseconds on the Jetson Nano. Subsequently, after obtaining the probability vectors of the network being employed, the actual position has to be inferred. Finally, this information is transformed into the FEN notation. The sum of the time required to execute these two operations is slightly over $10$ milliseconds. Therefore, roughly $100$ milliseconds must be added to complete the digitization.

Table \ref{tab:resumen_tiempo_total} summarizes the execution times of the aforementioned tasks when performing the whole process. As can be observed, when employing SqueezeNet-v1.1 or MobileNetV2($\alpha=0.5$), it is possible to digitize an image in times that range from $3.8$s to $5$s.

\begin{table}[]
    \centering
    \resizebox{\textwidth}{!}{%
        \begin{tabular}{@{}llccccc@{}}
            \toprule
            \multicolumn{2}{l}{}                                  & Test 1  & Test 2  & Test 3  & Test 4  & Test 5 \, \\ \midrule
            \multicolumn{2}{l}{Board detection}                 & $4.21$s & $3.30$s & $4.28$s & $3.69$s & $3.35$s \\ \midrule
            \multicolumn{2}{l}{\begin{tabular}[c]{@{}l@{}}Separate individual\\ squares\end{tabular}}                     & \multicolumn{5}{c}{$0.08$s} \\ \midrule
            \multirow{4}{*}{\begin{tabular}[c]{@{}l@{}}\, Obtain probability\\\,  vectors\end{tabular}} & SqueezeNet-v1.1 & \multicolumn{5}{c}{$0.46$s} \\
            & MobileNetV2 ($\alpha = 0.5$) & \multicolumn{5}{c}{$0.60$s}                     \\
            & NASNetMobile                 & \multicolumn{5}{c}{$3.42$s}                     \\
            & Xception                     & \multicolumn{5}{c}{$5.61$s}                     \\ \midrule
            \begin{tabular}[c]{@{}l@{}}\, Infer pieces +\\\, FEN notation\end{tabular}                       &                 & \multicolumn{5}{c}{$0.01$s} \\ \bottomrule
            \multirow{4}{*}{\, Total} & SqueezeNet-v1.1              & $4.76$s & $3.85$s & $4.83$s & $4.24$s & $3.90$s \\
            & MobileNetV2 ($\alpha = 0.5$) & $4.90$s & $3.99$s & $4.97$s & $4.28$s & $4.04$s \\
            & NASNetMobile                 & $7.72$s & $6.81$s & $7.79$s & $7.20$s & $6.86$s \\
            & Xception                     & $9.91$s & $9.00$s & $9.98$s & $9.39$s & $9.05$s \\ \bottomrule
        \end{tabular}
    }
    \caption{Summary of the total times on the Jetson Nano for each test board and for the models that form the Pareto front.}
    \label{tab:resumen_tiempo_total}
\end{table}

In situations where the camera and the board remain still, we have introduced in Section \ref{subsec:LC2FEN_board_detection} an algorithm to check if the board's location in the image is the same as in previous images. Running this algorithm on the Jetson Nano takes about $150$ milliseconds per board. Therefore, when this test returns a positive result, a huge reduction in the total execution time is achieved. In Table \ref{tab:resumen_tiempo_continuacion} the times when this board check returns true are summarized. The slight extra cost added when the test is false is highly compensated. In Table \ref{tab:resumen_tiempo_total} it is shown that the board that is detected the fastest takes $3.30$ seconds. Therefore, if this check returned true 1 out of 14 times, these calls would be amortized.

\begin{table}[]
    \centering
    \begin{tabular}{@{}llc}
        \toprule
        \multicolumn{2}{l}{}                                  & Static board and camera  \\ \midrule
        \multicolumn{2}{l}{\begin{tabular}[c]{@{}l@{}}Board\\ check\end{tabular}}                 & $0.15$s \\ \midrule
        \multicolumn{2}{l}{\begin{tabular}[c]{@{}l@{}}Separate individual\\ squares\end{tabular}}                     & $0.08$s \\ \midrule
        \multirow{4}{*}{\begin{tabular}[c]{@{}l@{}} \, Obtain probability\\\, vectors\end{tabular}} & SqueezeNet-v1.1 & $0.46$s \\
        & MobileNetV2 ($\alpha = 0.5$) & $0.60$s                     \\
        & NASNetMobile                 & $3.42$s                     \\
        & Xception                     & $5.61$s                     \\ \midrule
        \begin{tabular}[c]{@{}l@{}}\, Infer pieces +\\\, FEN notation \end{tabular}                       &                 & $0.01$s \\ \bottomrule
        \multirow{4}{*}{\, Total} & SqueezeNet-v1.1              & $0.70$s \\
        & MobileNetV2 ($\alpha = 0.5$) & $0.84$s \\
        & NASNetMobile                 & $3.66$s \\
        & Xception                     & $5.85$s \\ \bottomrule
    \end{tabular}

    \caption{Summary of the total times on the Jetson Nano for each test board and for the models that form the Pareto front when the board location check returns true.}
    \label{tab:resumen_tiempo_continuacion}
\end{table}

As can be seen, this prediction would reduce the total time required to digitize new positions to less than one second. In addition, periodic sampling can be added to capture possible intermediate moves. In this way, cases in which there is, for example, a hand covering part of the board could be avoided. Moreover, sometimes players forget to press the clock. In these cases, this periodic sampling would be necessary in order to capture all the moves.

\section{Conclusions} \label{sec:conclusions}

In this paper we have presented LiveChess2FEN, a framework for categorizing chess pieces employing a low-cost and low-power Nvidia Jetson Nano embedded device. Leveraging this device's parallelization capabilities, we have entirely digitized an image in less than 1s without the need for connectivity to any network, achieving around a 5$\times$ speedup over the state-of-the-art techniques while maintaining the same accuracy level.
To the best of our knowledge, this is the first attempt at deploying a chess digitization framework onto an embedded platform, obtaining similar if not better execution times than other approaches, which at least used a mid-range laptop to perform the tests.

Regarding accuracy when classifying the pieces, the investigated method obtains comparable results to those of other proposals.  Several CNNs have been tested, reaching a good trade-off between speed and accuracy with SqueezeNet and MobileNetV2. On top of this, several domain-based rules have been considered to optimize accuracy further. In this manner, we have avoided the usage of CPU intensive chess engines as in the literature solutions.

Notably, the training and testing in our experiments have been performed with different chess sets, which highlights the robustness that can be achieved using CNNs.  The complete source code of the LiveChess2FEN framework is publicly available with an open-source license in our GitHub repository: \url{https://github.com/davidmallasen/LiveChess2FEN}.

Finally, we should note that we have found a barrier to improving the piece classification accuracy while training different CNNs. The simplest models have been able to learn practically all the information available in our dataset, and training more complex models has provided almost no benefit (Table \ref{tab:jetson_top1vsdominio}). With this in mind, we believe that part of this problem would be solved by introducing more information into the dataset.

\subsection*{Acknowledgements}

This paper has been supported by the CM under grant S2018/TCS-4423, the EU (FEDER) and the Spanish MINECO under grant RTI2018-093684-B-I00 and by Fundación BBVA under grant PR2003\_20/01.

\bibliographystyle{acm}
\bibliography{LiveChess2FEN}

\end{document}